# Facial 3D Model Registration Under Occlusions With SensiblePoints-based Reinforced Hypothesis Refinement

Yuhang Wu and Ioannis A. Kakadiaris
Computational Biomedicine Lab, University of Houston
{ywu35,ikakadia}@central.uh.edu

## Abstract

*Registering a 3D facial model to a 2D image under occlusion is difficult. First, not all of the detected facial landmarks are accurate under occlusions. Second, the number of reliable landmarks may not be enough to constrain the problem. We propose a method to synthesize additional points (SensiblePoints) to create pose hypotheses. The visual clues extracted from the fiducial points, non-fiducial points, and facial contour are jointly employed to verify the hypotheses. We define a reward function to measure whether the projected dense 3D model is well-aligned with the confidence maps generated by two fully convolutional networks, and use the function to train recurrent policy networks to move the SensiblePoints. The same reward function is employed in testing to select the best hypothesis from a candidate pool of hypotheses. Experimentation demonstrates that the proposed approach is very promising in solving the facial model registration problem under occlusion.*

## 1. Introduction

The goal of facial model registration (pose estimation) is to estimate a projection matrix **P** that is able to fit a dense 3D facial model $\mathbb{Y}_{3D}$ onto a 2D facial image where the point-wise correspondences between the 3D model and the 2D image are well constructed. This module is heavily used in normalizing [7, 11, 16, 31] or rendering [12, 24, 25, 34] facial textures, which are critical preprocessing stages to reduce pose variations in face recognition (FR). With texture transformation, newly proposed approaches [24, 35] have been demonstrated to be more robust to pose variations than common deep networks using identical or less training data. However, the robustness of facial model registration limits the use of FR in-the-wild (*e.g.*, under partial occlusions, extreme head pose variations in surveillance view). In these challenging scenarios, failures of model registration are mainly caused by external occlusions (*e.g.*, hair, mask, glasses, and shade, etc.) and out-of-plane head rotations (internal occlusions). The proposed method is mainly developed for improving the robustness of pose-invariant FR [7, 11, 12, 16, 24, 25, 31, 34] in more unconstrained scenarios, especially under internal or external occlusions.

Accurate 3D registration under occlusion is a difficult problem, even using a deep neural network. Small errors in pose parameter estimation can easily cause global misalignment. Hence, a capability to rectify alignment error on-line is needed to solve this problem. We define a reward function that a system can access after deployment (in testing). This reward function has a dual use. In training, it helps to train a recurrent neural network for model registration. In testing, it helps to select the best pose hypothesis. Through this approach, the model has the capability to measure alignment error after deployment, which contributes to improving the alignment accuracy under occlusion.

The proposed method is built upon a relaxed assumption on landmark (facial key-points) detection. Instead of assuming all fiducial points (landmarks have clear semantic definitions *e.g.*, eye corners) can be accurately detected, we assume that only a portion of fiducial points (FPs) can be reliably annotated and propose pose hypotheses. This assumption can be easily satisfied with a state-of-the-art landmark detector, where visible landmarks have a much higher chance to be annotated accurately. However, this assumption brings a problem that the number of reliable landmarks are significantly reduced under occlusions, which yields a weakly-constrained pose estimation problem. To resolve this problem, we impose a secondary assumption on non-fiducial points (NFPs, landmarks without clear semantic definitions) on the facial contour (CL). We assume that they are able to provide visual clues to help verify whether the silhouette of a 3D model is accurately registered, even though they themselves are not reliable enough to propose pose hypotheses. This assumption gives us a possibility to evaluate the performance of model registration by passively checking whether the NFPs are well-aligned, and finally helps us to constrain the problem.

The main contribution of the proposed method is that we convert the traditional 3D-to-2D pose estimation problem defined on the point-cloud space into a search problem solved in confidence-map space. The pipeline of the proposed method is shown in Fig. 1. We propose a method to synthesize new 2D points automatically for pose hypothesis generation; we call the newly synthesized landmark 'SensiblePoints' (SPs). We first initialize the locations of $N$ SPs based on a newly proposed retrieval technique. Then, $N$ recurrent neural networks are trained to move the $N$ SPs (one for each) on the 2D image to maximize the aforementioned reward function using the REINFORCE algorithm [32]. New pose hypotheses are generated based on the new locations of SPs and the visible FPs. Based on all proposed hypotheses, we seek the hypothesis that is best able to explain all the visible clues contributed by the confidence-maps in testing. To the best of our knowledge, this is the first method that uses deep reinforcement learning for 3D-to-2D facial model registration under occlusion. The main contributions of our work include:

- We propose a robust facial model registration method that can exploit the information encoded in the confidence-maps generated by fully convolutional networks.

- We propose a new method that uses SensiblePoints to generate pose hypothesis and constrain model registration.

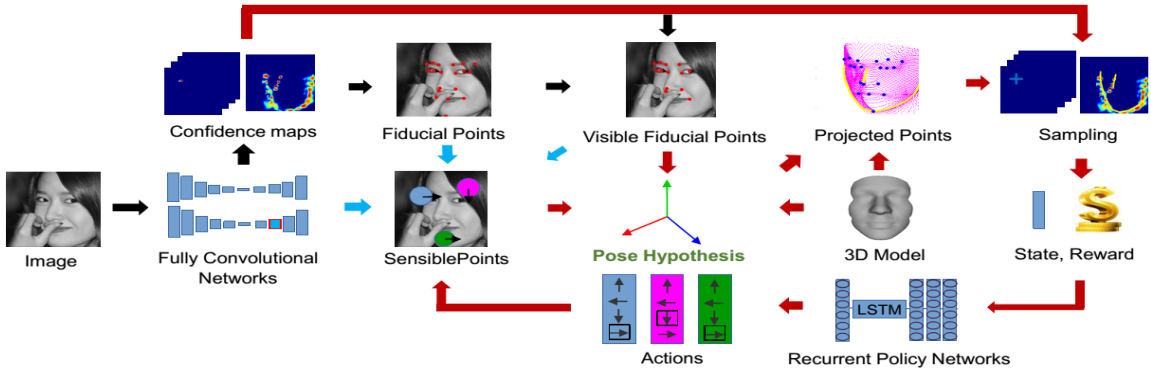

Figure 1: Overview of the proposed approach.

## 2. Related work

Existing works in facial model registration can be divided into two categories: landmark-based registration and landmark-free registration. In the first category of methods, 2D coordinates of FPs (denotated as $\mathbf{x}_a$) are first estimated using an automatic detector [3, 35, 21, 2, 28, 37, 39], then a projection matrix $\mathbf{P}$ is estimated to minimize a reprojection error between $\mathbf{x}_a$ and the projected 2D locations computed by $\mathbf{P} \cdot \mathbf{y}_a$, where $\mathbf{y}_a$ is a vector containing the coordinates of corresponding 3D fiducial points on $\mathbb{Y}_{3D}$. Weak-prospective projection model, golden standard approaches [10], POSIT [5], and other approaches [6] can be used to estimate the $\mathbf{P}$ based on $\mathbf{x}_a$ and $\mathbf{y}_a$. Because this category of methods assumes the facial landmarks are detected, their performance relies heavily on automatic landmark detectors. In landmark-free model registration [41, 15, 36], a 3D model is directly registered to the 2D image based on the feature extracted from the facial ROI. Because the losses are enforced on the pose parameters, this category of methods is not quite aware of local misalignment of landmarks.

Current state-of-the-art approaches for fiducial points (FPs) detection are based on confidence-maps generated by fully convolutional networks [3, 35, 21, 2]. In these works, a confidence-map is generated for each landmark to indicate the possibility of a landmark appearing at a specific location in the original image. The methods are highly accurate in localizing visible landmarks. In this paper, we develop a classifier to discriminate the visible and occluded landmarks based on the confidence-maps, and only use the visible landmarks for proposing a hypothesis. Unlike the FPs that are able to provide a valid pose hypothesis, non-fiducial points (NFPs) provide important clues to measure the performance of registration. Previous approaches [39, 28, 37] are able to detect 17 NFPs on 2D facial contour. However, they are not designed for 3D registration, since their locations do not correspond to the same semantic positions on the 3D facial model when the facial pose varies [41]. In our approach, the confidence-maps of NFPs on the facial contours contribute to generate a reward score that measures the alignment between the silhouette of the projected 3D model and the facial contour on the image. We noted that recently a few works attempt to directly detect the points on 3D facial contours [3, 2]. In our work, we do not employ the points on 3D facial contour for hypothesis proposal because we observed that these points are relatively more sensitive to bounding box initialization and external occlusions. We argue that for model registration it is more robust to make full use of the visual clues provided by the confidence-maps instead of relying on deterministic prediction on occluded landmarks.

To improve the robustness of model registration, a few approaches that make their decisions based on hypotheses evaluation. RANSAC [8] is a classical approach in this category. However, this method requires a large pool of point-wise correspondences, which is infeasible in our problem. Recent approaches [19] score pose hypotheses by comparing rendered and observed image patches and obtain impressive results in object pose estimation. Krull *et al.* [20] used a policy network to refine pose hypotheses. Compared with [19, 20], 3D-to-2D projection is challenging because we do not have depth information to generate rewards or verify pose hypotheses. Instead, the hypothesis is proposed by moving SPs instead of selecting a subset from a 3D landmark pool as in [20]; in addition, rewards are measured on the 2D confidence-maps instead of on the pose parameter space.

## 3. Method

### 3.1. Landmark detection under occlusion

**FP-net - detecting fiducial landmarks:** To obtain the $U$ fiducial landmark locations $\mathbf{x}_a$, where $\mathbf{x}_a$ is a vector that records the two dimensional coordinates of all FPs, a fully convolutional network [35] is employed. This architecture achieved state-of-the-art results in localizing visible landmarks under out-of-plane head rotation. For localizing occluded landmarks, this approach is forced to make a prediction even though it is not confident about a landmark location, which yields higher localization error. We use $\mathbb{H}_a$ (80×80×19 dimensions) to denote the confidence-maps generated by FP-net.

**NFP-net - detecting non-fiducial landmarks:** To detect $V$ non-fiducial points on the facial contours, a separate, fully convolutional network is trained whose output confidence-maps are denoted as $\mathbb{H}_b$ (80×80×21 dimensions), and whose landmark vector is denoted as $\mathbf{x}_b$. A one-level hour-glass network introduced in [35] is employed as the network architecture. We did not jointly train a network to detect both FPs and NFPs because we want the network to be able to make an objective observation as to where the NFPs are instead of relying on privileged information of fiducial landmarks to make an inference; this information might not be reliable under external occlusions.

### 3.2. Landmark visibility inference

To estimate the landmark visibility, we propose a new approach. Although more complex approaches such as deep neural networks [29], surface normal [15], or deformable part model [9] can be employed, we propose to employ a light-weight, general-purpose module that is built upon the confidence-maps of the FP-net. As shown in Fig. 2(L), the confidence-map of a visible landmark has the following pattern: (1) high contrast between foreground (landmark) and background regions, (2) small but concrete foreground region. In contrast, as shown in Fig. 2(R), the confidence-map of an occluded landmark has a lower contrast but larger response region, with a histogram showing a long-tail distribution. To model this observation, two features are extracted from a confidence-map.

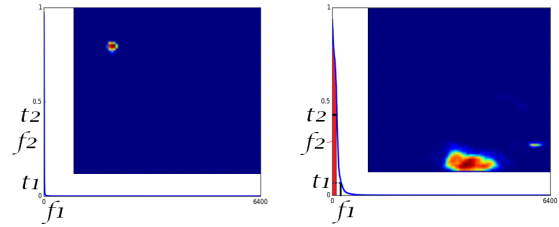

Figure 2: (L) Confidence-map of a visible fiducial point, and (R) Confidence-map of an occluded fiducial point. The pixel values in the confidence-maps are sorted and the corresponding histograms are shown below. The thresholds and features extracted from the histograms are highlighted.

**Maximum likelihood classifier:** First, the percentage of pixels with confidence (pixel value) beyond an adaptive threshold $t^1$ is characterized. The highest response value in a confidence-map is extracted, denoted by $\theta_a$, then a threshold $t^1$ is computed by $\epsilon_1 \times \theta_a$. The percentage of pixels with values larger than $t^1$ are computed in the confidence-map and this is the first feature $f^1$. The reason that the relative threshold is employed instead of an absolute threshold is that we observed $\theta_a$ is changing due to the variation of image resolution. To make it more robust to this variation, a relative threshold is a better choice. However, using only this threshold is not enough. Consider the case of a confidence-map with a histogram that follows a long-tail distribution due to some external occlusions, but also has a very sharp peak which marks the visible landmark location. Obviously, $f^1$ will not capture the sharp peak and focuses on the long-tail distribution. To overcome this disadvantage, a second feature is proposed. The sum of pixel values with confidence above a pre-determined threshold $t^2$ is computed, and denoted as $u^1$ ($t^2$ is learned by a line-search in training.) The second feature $f^2$ is computed as $u^1/u^2$, where $u^2$ is the summation of all pixel values inside the network. The letter $f^2$ focuses on the contributions of high confidence regions instead of the long-tail in the histogram. A binary maximum-likelihood classifier is trained on these two features to predict the landmark visibility.

**Detector-specific visibilty inference:** In practice, we observe that even though the maximum likelihood classifier predicts that a landmark is occluded, the FP-net may still localize it accurately (*e.g.*, locations of eyes behind sun glasses). This problem requires adjusting the decision boundary of the two classes and determining the 'visibility' from a perspective that takes into account the landmark localization error of FP-net. In our implementation, only if the probability of predicting a landmark is 'invisible' (by the maximum-likelihood classifier) that is $\zeta$ times larger than the probability of predicting the landmark is 'visible', the landmark is marked as truly 'invisible' (unable to annotate correctly by FP-net). In the training set, $\phi_v$ is used to denote the average normalized mean error (NME) [41] of a landmark whenever it is predicted as visible, $\phi_{\tilde{v}}$ to denote the average normalized mean error of the landmark whenever it is predicted as occluded. The scalar $\zeta$ is a value that maximizes $\phi_{\tilde{v}}$-$\phi_v$. Notice that each landmark has its own $\zeta$, and their values are learned by line-search on the training set.

Now we are capable of selecting the coordinates of visible (reliable) landmarks from $\mathbf{x}_a$. The vector of visible FP is denotated as $\mathbf{x}_v$, $\mathbf{x}_v \subseteq \mathbf{x}_a$.

### 3.3. SensiblePoints and initialization

After estimating the visibility of landmarks, a small portion of reliable landmarks are obtained.

**SensiblePoints:** SensiblePoints (SPs) are defined on a few FPs with the probability of being occluded due to out-of-plane rotations or external occlusions (*e.g.*, hair, masks, and glasses) is higher than the others. The letter $N$ is used to denote the number of SPs. In our paper, $N$ is assigned to be three. Figure 3 shows their initial locations. SPs are used to replace the corresponding FPs in model registration (the corresponding positions on the 3D model are not changed). The locations of SPs are not constrained by the initial locations of FPs; they can move freely inside an image to correct mis-alignment of registration caused by FPs. The goal of SPs is to behave as visible FPs, work along with $\mathbf{x}_v$, and propose pose hypothesis. We call it 'sensible' because these points are able to explore image space and generate good pose hypotheses by identifying the pattern of local mis-alignment.

**Robust shape retrieval:** We hope to initialize the SPs close to their final positions to speed up convergence. This initialization approach should be able to exploit the knowledge encoded in the FP-net and also be robust to out-of-plane rotations (internal occlusion). We propose a new approach that uses the contextual information extracted from the neighborhood patches of FPs to retrieve the target shape. The method is based on the local volumes cropped from an intermediate layer of the FP-net. Because FP-net is a fully convolutional network, an FP location in the output layer of the network can be mapped back into a corresponding location in any intermediate layer of FP-net with a scaling factor $q$. In practice, $M$ feature volumes are extracted from an intermediate network layer, which corresponds to $M$ neighborhood patches of FPs in the output layer (as shown in Fig. 3(a)). The response region of each volume covers $1/4^2$ of the output confidence-maps. The $M$ FP positions are selected close to the symmetric axis of the face that can nicely track the pose of faces under out-of-plane head rotations. Later, the $M$ feature volumes are unraveled and concatenated into a 1D feature vector $\mathbf{f}_M$ as the current representation of the estimated facial shape. Five images from a reference database are retrieved with feature vectors that are closest to $\mathbf{f}_M$ in the Euclidean space. The shapes of the retrieved images are denoted as $\mathbf{x}_{a'}^i$, ($i$=1,2...5). Notice that the $\mathbf{f}_M$ is extracted from the intermediate layer regardless the visibility of the corresponding $M$ FPs. The contextual information in the occluded patterns (*e.g.*, glasses, mask, etc.) and facial pose help to retrieve similar facial shapes from the reference database because they are encoded in the intermediated layers of the network.

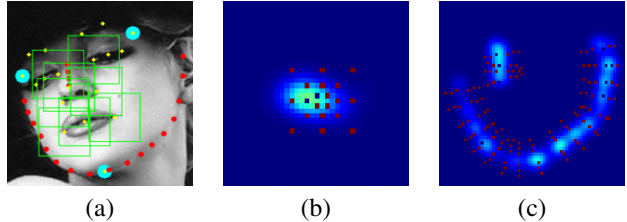

Figure 3: (a) Depiction of the Landmark layouts. The yellow dots depict the locations of 19 FPs, the red dots depict the locations of 21 NFPs, the blue rounds depict the locations of SPs, green rectangles depict the response regions of $\mathbf{f}_M$. Notice that there are two overlapping points in NFPs and FPs: 'nose tip' and 'chin center' share the same semantic meaning on the face. These two points in NFPs are employed because they help us capture the silhouette outline of the face. (b) Binary descriptor on a confidence-map of $\mathbb{H}_a$, blue dots depict one, red dots depict zero. (c) Binary descriptor on $\mathbb{H}_b^\Sigma$, blue dots depict one, red dots depict zero.

**Intermediate network layer selection:** Since the Euclidean distance between $\mathbf{f}_M$ is computed in retrieval, the loss function of linear discriminant analysis (LDA) [23] is employed to select the best network layer in the FP-net for shape retrieval. Considering the time complexity, only a layer down-sampled eight times or more is employed in the selection. The pose spaces in the reference database are first divided into 45 classes (five yaw variations, three pitch variations, and three roll variations), the layer that is able to generate an $\mathbf{f}_M$ that can best discriminate the 45 classes using the LDA loss is employed. The $4^{th}$ deconvolutional layer in FP-Net is selected as the layer for shape retrieval.

**SensiblePoints initialization:** First, the head pose is estimated based on $\mathbf{x}_v$ and $\mathbf{y}_v$. The vector $\mathbf{y}_v$ contains corresponding 3D locations of the 2D visible points on the 3D model. When the head pose is near frontal, the $\mathbf{x}_v$ is employed to estimate a 3D-to-2D projection matrix and the reprojected 2D points are used to initialize the SPs. When the head pose has a large deviation (yaw angle larger than

30°) from frontal, rigid 2D affine transformation is first employed to transform each retrieved shape $\mathbf{x}_{a'}^i$ to the original image based on the visible FPs $\mathbf{x}_v$. Then, a median shape is computed among the transformed shapes. Finally, the corresponding locations in the median shape are used to initialize the SPs.

### 3.4. Policy network for hypothesis refinement

**Hypothesis generation:** After initializing the SPs, a vector $\mathbf{x}_c$ is used to record the 2D coordinates of all SPs, then $\mathbf{x}_c$ is appended to $\mathbf{x}_v$ to obtain a vector $\mathbf{x}_h$: $\mathbf{x}_h = [\mathbf{x}_c; \mathbf{x}_v]$. A 3D-to-2D projection matrix $\mathbf{P}_h$ is estimated with $\mathbf{x}_h$ and its corresponding 3D landmark vector $\mathbf{y}_h$ on $\mathbb{Y}_{3D}$ based on the weak-perspective projection model, and the $\mathbf{P}_h$ is our pose hypotheses under $\mathbf{x}_h$. Compared to an estimate of $\mathbf{P}_h$ with only $\mathbf{x}_v$, $\mathbf{x}_h$ provides more constraints to the problem. Next, we introduce our method to move the SPs to generate better pose hypotheses. We name the method Sensiblepoints-based reinforced Hypothesis Refinement (SHR).

**Hypothesis updating:** Let $j$ represent the $j^{th}$ SP ($j \in \{1, 2, .., N\}$), $t$ represent the $t^{th}$ movement of the SP ($t \in \{1, 2, .., T\}$), and the hypothesis generated at the new location of SP $\mathbf{x}_c^{j,t}$ is denoted as $\mathbf{P}_h^{j,t}$. A specific deep recurrent network is trained to move each SP. Assume there is only one SP being moved at a time. The network updates the position $\mathbf{x}_c^{j,t}$ of the SP as: $\mathbf{x}_c^{j,t+1} = \mathbf{x}_c^{j,t} + \boldsymbol{\sigma}^{j,t}$, where $\boldsymbol{\sigma}^{j,t}$ is the output of the network. Unlike previous approaches that tried to minimize a re-projection error between $\mathbf{P}_h^{j,t} \cdot \mathbf{y}_h$ and $\mathbf{x}_h$, the network learns to move the SP to maximize an alignment score $s^{j,t}$ between the projected 2D points of the dense 3D model $\mathbb{Y}_{2D}^{j,t}$ ($\mathbb{Y}_{2D}^{j,t} = \mathbf{P}_h^{j,t} \mathbb{Y}_{3D}$) and the visual clues encoded in the confidence-maps: $\mathbb{H}_a$ and $\mathbb{H}_b$. Because this score passively samples the pixels from $\mathbb{H}_a$ and $\mathbb{H}_b$ based on the locations of FPs and facial contours, the number of sampled pixels varies under different head pose and occlusions, it is not differentiable w.r.t. $\boldsymbol{\sigma}^{j,t}$ or $\mathbf{P}_h^{j,t}$. To solve this problem, the policy gradient approach [32] in deep reinforcement learning is employed to optimize parameters inside the network.

**Recurrent policy network:** The recurrent network is the engine of SP. The input of the recurrent network is a state vector denoted as $\boldsymbol{\tau}^{j,t}$; it is used to describe the status of $\mathbb{Y}_{2D}$ on $\mathbb{H}_a$ and $\mathbb{H}_b$. The output of the network is an action vector denoted as $\boldsymbol{\sigma}^{j,t}$; it is used to update the location of an SP. The reward signal, denoted as $r^{j,t}$, plays a critical role in optimizing the network parameters, and represents the increment of alignment score before and after moving the SP. The recurrent network learns a mapping $\pi^{j,t}$ from $\boldsymbol{\tau}^{j,t}$ to $\boldsymbol{\sigma}^{j,t}$: $\boldsymbol{\sigma}^{j,t} = \pi^{j,t}(\boldsymbol{\tau}^{j,t})$, to maximize the expected reward in all the recurrent stages. This mapping is called 'policy' in reinforcement learning, and our recurrent network can also be called a 'policy network'. As shown in Fig. 1, our policy network contains three fully connected layers, an LSTM unit [13], and an output layer that transforms the hidden layer in LSTM to $\boldsymbol{\sigma}^{j,t}$. Each network layer (including the hidden layer of LSTM) contains 64 hidden neurons. The Leaky-ReLU [22] is employed as the activation function. To avoid gradient explosion, the gradients clipping [26] is employed. ADAM [17] is used to update the parameters inside the network.

**Action:** To reduce the variance of estimated gradient in training and speed up convergence, a small $\boldsymbol{\sigma}^{j,t}$ space is employed which contains four possible actions (classes): move the $j^{th}$ SP 'up', 'down', 'left', or 'right' by $\epsilon_2 \cdot \omega$ pixels in the original image space, where $\omega$ is the square root of face ROI. The policy network selects an action to take from the four candidate actions according to a soft-max layer.

**Rewards:** The reward $r^{j,t}$ under $t^{th}$ recurrent stage is computed as: $r^{j,t} = s^{j,t} - s^{j,0}$, where $s^{j,0}$ is the alignment score generated by the initial pose hypothesis, which is computed by the FPs with the initialized SPs.

**Alignment Score:** The $s^{j,t}$ is computed by checking whether $\mathbb{Y}_{2D}$ is well-aligned with the high confidence regions on $\mathbb{H}_a$ and $\mathbb{H}_b$. An alignment score $s^{j,t}$ is computed by:

$$s^{j,t} = \lambda_1 s_1^{j,t} + \lambda_2 s_2^{j,t} + \lambda_3 s_3^{j,t} \quad (1)$$

where $s_1^{j,t}$ measures the alignment accuracy of (projected) visible FPs on $\mathbb{H}_a$, $s_2^{j,t}$ measures the alignment accuracy of (projected) occluded FPs on $\mathbb{H}_a$ ($s_2^{j,t}$ is much less than $s_1^{j,t}$), and $s_3^{j,t}$ measures the alignment accuracy of the silhouette outline of $\mathbb{Y}_{2D}$ on $\mathbb{H}_b$. $\lambda_1$, $\lambda_2$, and $\lambda_3$ are weights, where $\lambda_1 + \lambda_2 + \lambda_3 = 1$. The upper bound of $s^{j,t}$ may vary due to different occluded conditions. To compute $s^{j,t}$, all confidence-maps in $\mathbb{H}_a$ and $\mathbb{H}_b$ are blurred to increase the tolerance to slight alignment error. To compute $s_1^{j,t}$, $\mathbb{H}_a$ is sampled using: $\mathbf{P}_h^{j,t} \cdot \mathbf{y}_v$ – the projected positions of visible fiducial points (FP) in $\mathbb{Y}_{2D}$. When sampling the $\mathbb{H}_a$, each projected FP picks a pixel value from its own confidence-map regardless of the others. The sampled pixel values on the confidence-maps are averaged to be $s_1^{j,t}$. $s_2^{j,t}$ is computed based on the confidence-maps of occluded landmarks, where the pixel locations at $\mathbf{P}_h^{j,t} \cdot \mathbf{y}_{\tilde{v}}$ are sampled and averaged. The confidence-maps of occluded landmarks are usually employed to generate prediction directly, this deterministic strategy is avoided by softly aggregated confidence values and weighted by a $\lambda_2$. $s_3^{j,t}$ are sampled with $\mathbf{P}_h^{j,t} \cdot \mathbf{y}_b$, based on $\mathbb{H}_b$. A similar approach as [42] is employed to obtain the projected positions of NFPs on $\mathbb{Y}_{2D}$. We observed that the projected NFPs from the 3D model are not guaranteed to be well-aligned with the confidence maps generated by NFP-net, even if the model is well registered. This may cause consistent offsets and reduce alignment score. In our solution, the projected NFPs on $\mathbb{Y}_{2D}$ are connected to create two sampling lines $\Theta_1$ and $\Theta_2$. One for sampling the confidences of NFPs on facial contour and one for sampling the NFPs on the nose. These two lines are used to sample a confidence-map that sums up all confidence-maps in $\mathbb{H}_b$, denoted as $\mathbb{H}_b^\Sigma$ and the average pixel values is $s_3^{j,t}$.

**State:** The state $\tau^{j,t}$ (input) in our network is a 1D descriptor that senses the alignment of the projected locations of $\mathbf{P}_h^{j,t} \cdot \mathbf{y}_a$ and $\mathbf{P}_h^{j,t} \cdot \mathbf{y}_b$ on the high confidence regions of $\mathbb{H}_a$ and $\mathbb{H}_b^\Sigma$. Because confidence-maps are highly discriminative, features are directly extracted from this space without using another deep neural network. A 20-bit binary descriptor covering a 14×14 neighborhood pattern around each projected location of FP is employed to describe the alignment of FP. The pattern of this descriptor is shown in Fig. 3(b). Each bit in the descriptor is computed by comparing the value of the center pixel with another sampling pixel in the descriptor sequentially. If the value of the center pixel is larger than the other sampling pixel, zero is appended to the binary descriptor, otherwise one is appended. Therefore, if a projected FP is well-aligned with its corresponding confidence-map, all 20 bits are zero. Thus, this descriptor will contribute zero in moving the SP. If a landmark is occluded, an all-zero 20-bit descriptor is also employed, because the information extracted from the confidence-map is not reliable enough to move the SP. For NFPs, a descriptor is used to measure the alignment between $(\Theta_1, \Theta_2)$ and $\mathbb{H}_b^\Sigma$. This alignment is characterized using a descriptor defined on $\mathbf{P}_h^{j,t} \cdot \mathbf{y}_b$. For each projected NFP, a six-bit descriptor is extracted on the two sides of the NFP as shown in Fig. 3(c), the sampling positions are perpendicular to $\Theta_1$ or $\Theta_2$. The distance between any two sampling positions is controlled to be two pixels, this value may vary up to one pixel depending on the local gradients of $\Theta_1$ or $\Theta_2$. Finally, a $(U \times 20 + V \times 6)$ dimension feature vector is obtained to describe 3D model alignment. Inspired by [33], the descriptor is concatenated with a vector, which is computed using the z-score of $\mathbf{P}_h^{j,t} \cdot \mathbf{y}_a$, to better encode the current facial pose, and non-rigid information of the face. After concatenating, the final descriptor, with the dimension $(U \times 20 + V \times 6 + U \times 2)$, is our state vector: $\tau^{j,t}$.

---

**Algorithm 1** Sensiblepoints-based Hypothesis Refinement

**Input:** A facial image, a 3D dense model $\mathbb{Y}_{3D}$
**Output:** 3D-to-2D projection matrix $\mathbf{P}$
**Step 1:** Predict $\mathbf{x}_a$ via fully convolutional network
**Step 2:** Obtain confidence maps $\mathbb{H}_a$, $\mathbb{H}_b$, compute $\mathbb{H}_b^\Sigma$
**Step 3:** Estimate visibility of FPs and obtain $\mathbf{x}_v$
**Step 4:** Initialize $\mathbf{x}_c$ of SPs, let $\mathbf{x}_h = [\mathbf{x}_c; \mathbf{x}_v]$
**Step 5:** Propose and refine pose hypothesis by updating $\mathbf{x}_c$ in $\mathbf{x}_h$ based on recurrent policy networks
**Step 6:** Select the best pose hypothesis that yields the highest alignment score on $\mathbb{H}_a$ and $\mathbb{H}_b^\Sigma$ to be $\mathbf{P}$

---

## 4. Inference

In testing, the locations and visibilities of FPs are estimated, then SPs are initialized. The policy networks are employed to update $\mathbf{x}_c^0$ in an interleaved manner to generate pose hypotheses. When a policy network moves an SP, the locations of other SPs are kept unchanged as the initial value. After the first iteration, the policy networks move all the $N$ SPs and generate $N \times T_1$ pose. The SP locations that yield the highest alignment score are used to initialize the SPs in the next iteration (hypothesis refinement). So in each iteration, only one SP is moved. The system is stopped after $T_2$ iterations, the hypothesis that yields the highest alignment score among all the iterations is selected to be the final hypothesis. The summary of the proposed algorithm is shown in Alg. 1.

## 5. Experiment

**Databases:** The database partition that is employed in training and testing is shown in Table 1.

Table 1: Database partition. (*) When NME/NRME is computed, any overlapped points between NFP and FP are counted once. So it should be 19+21-2=38 landmarks employed for evaluation.

| Module | Usage | Database | Imgs. | Ldmks. |
|---|---|---|---|---|
| P-Net | Train | AFLW-Tr | 20,000 | 19(FP) |
| NFP-Net | Train | H/L-LP-Tr | 18,036 | 21(NFP) |
| Visibility | Train | COFW-Tr | 1,345 | 19(FP) |
| Policy-Net | Train | COFW-Tr | 1,345 | 19(FP) |
| Policy-Net | Vali. | H/L-Te | 630 | 38* |
| SP-Init. | Ref. | H/L-LP-Te | 2,000 | 19(FP) |
| All | Test | iBUG-135 | 135×6 | 38* |
| All | Test | COFW-Te | 507 | 19(FP) |

The training databases of AFLW-Full [40, 18], 300W-LP [41, 30] (HELEN set [38] and LFPW set [1]), and COFW [4] are employed for learning; the original testing set of HELEN, LFPW (without 3D augmentation) for parameter selection (validation), and 2,000 random selected images from the testing set of 300W-LW (HELEN and LFPW) for shape retrieval; the iBUG challenging set (iBUG-135) [30] and the testing set of COFW are used for evaluation. Random occlusion patches are added on iBUG-135 to evaluate the proposed method under occlusions. As shown in Fig. 4, the occlusion patches cover 1%, 4%, 9%, 16%, 25% of face ROI. For 3D models, the parameters provided by [41] are employed to reconstruct personalized 3D BFM models [27] for iBUG-135; a generic BFM model is employed for model registration on COFW database.

**Error measurement:** Aside from the normalized mean error (NME), an error called normalized re-projection mean error (NRME) is employed to evaluate the 3D model registration error. This error measures how well the 3D model is aligned with the 2D image:

$$NRME = \frac{1}{N' \times \omega} \sum_{j}^{N'} ||\mathbf{P} \cdot \mathbf{y}^j - \mathbf{x}_{GT}^j||,$$

where $\mathbf{P} \cdot \mathbf{y}^j$ is the 2D projected positions of $N'$ 3D landmarks under pose hypothesis $\mathbf{P}$, $\mathbf{x}_{GT}^j$ is the ground-truth 2D landmarks, the square root of face ROI is $\omega$. To compute NRME, $\mathbf{P}$ is first computed based on FPs (plus SPs) with weak-perspective projection model, then NRME is computed by using $\mathbf{P}$ and $\mathbf{y}^j$. As in [35], the function provided by the AFLW database is employed to initialize the ROI.

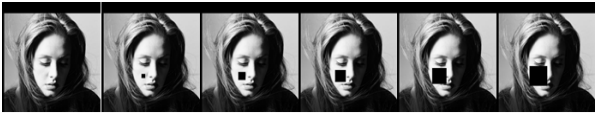

Figure 4: iBUG-135 database with synthesized occlusions.

**Parameter selection:** The COFW training set is used to learn the model parameters for visibility inference. To determine the $\lambda$s in Equation (2), the $\lambda_1$, $\lambda_2$, and $\lambda_3$ are first initialized to be 0.3, 0.2, and 0.5 empirically. The recurrent policy networks are trained for three epochs on COFW training set. Then, let $\lambda_2 = 0$ and an optimal aspect ratio of $\lambda_1 \backslash \lambda_3$ is computed on the validation set as follows: for each image, the inference algorithm is employed (Section 4) and the values of $s_1^{j,t}$, $s_3^{j,t}$, and the NRME in each recurrent stage are saved. Based on these saved values, a $\lambda_1^*$ that can best minimize the rank loss (Normalized Discounted Cumulative Gain [14]) between the NRMEs and the $\lambda_1 s_1^{j,t} + \lambda_3 s_3^{j,t}$ ($t = 1, 2, ...T_1 \cdot T_2$) is selected. In our solution, the $\lambda_1^*$ is determined by a line search in [0:0.01:1]. After obtaining the $\lambda_1^*$ for each individual image, a mean $\lambda_1^*$ (denoted as $\bar{\lambda}_1^*$) is computed over all images in the reference database and the final $\bar{\lambda}_1^*\backslash\bar{\lambda}_3^*$ is obtained (notice that $\lambda_3^* = 1 - \lambda_1^*$, $\lambda_2 = 0$). Then, the $\bar{\lambda}_1^*\backslash\bar{\lambda}_3^*$ is fixed and the same approach is employed to determine the ratio of $(\bar{\lambda}_1^* + \bar{\lambda}_3^*)\backslash\bar{\lambda}_2^*$ on the COFW database. Finally, the constraints of $\bar{\lambda}_1^* + \bar{\lambda}_2^* + \bar{\lambda}_3^* = 1$ are used to determine the specific value of each weight. The final values of the $\lambda_1, \lambda_2$, and $\lambda_3$ are 0.21, 0.60, and 0.19. These new parameters are used to retrain the recurrent policy networks for three epochs on the COFW training database to obtain the final model. The hyperparameter $\epsilon_1$ is set to be 0.1 to capture the long-tail distribution of $\mathbb{H}_a$; $\epsilon_2$ is set to be 0.01 so that the SP will be gradually moved. $T_1$ and $T_2$ are set to be 30 and four, respectively, to guarantee the SPs sufficiently explore the image space. We did not observe further improvement with larger $T_1$ and $T_2$.

**Computing Time** It takes three days to train the recurrent policy network. In testing, the network takes 20 seconds to register a 3D model onto a 2D image based on a GTX960 graphic card. Most of the time is spent computing the 2D projected positions of the Basel 3D model [27].

### 5.1. Module analysis

**SensiblePoints initialization under occlusion:** The objective of this experiment is to evaluate the robustness of the proposed SP initialization approach described in Section 3.3. The NME of the SPs are computed to measure the accuracy of initialization. The prediction made by FP-Net is compared. As shown in Fig. 5, the results demonstrate that the proposed SP initialization approach is more robust to image occlusion, which can provide better initial positions for the SPs under occlusions.

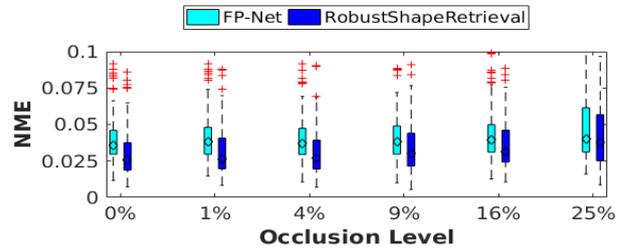

Figure 5: Boxplot of SensiblePoints initialization error on iBUG-135 database with synthesized occlusions.

**Policy network vs greedy policy:** The objective of this experiment is to evaluate the effectiveness of the policy network. The experiment is conducted on the iBUG-135 database without synthetic occlusion. In Fig. 6, the alignment score after registering a reconstructed 3D model to the 2D image is shown. The green line depicts the alignment score in Eq. 1 when the 3D model is registered by FPs. The red line depicts the alignment score after employing SHR. In the greedy policy, instead of using a recurrent policy network to select an action, a SensiblePoint greedily selects an action that yields the highest alignment score in the four possible actions. The same protocol introduced in Section 4 is employed in testing. The final alignment score is indicated as the blue line in Fig. 6. We observed that the recurrent policy network helps the SensiblePoints refine the initial pose hypothesis and outperforms the greedy policy.

### 5.2. Model registration under occlusion

**Register a reconstructed 3D model:** In [16, 24, 7, 11], a reconstructed 3D model is registered for FR based on FPs. In our work, the reconstructed 3D models are registered to the images in iBUG-135 database (with occlusions). FP(All) is used to denote the baseline that use all FPs generated by FP-Net to register a 3D model as [16, 24, 7, 11]. The FPs estimated by CFSS [39] are also used for registering 3D models for comparison. In the second baseline, the 3D model is registered by the visible FPs, which is denotated as FP(Vis). The proposed system is denoted as FP(Vis)+SHR. Because the ground-truth positions of FPs and NFPs are available, the NRME of 38 points (as shown in Fig. 3) are computed for comparison. As shown in Fig. 7 and Fig. 8, results indicate our model registration approach is robust to external occlusions. Compared to SHR, a recently proposed landmark-free model registration approach PAWF [15] is more sensitive to facial occlusions. The experiments demonstrate the SHR can improve the robustness of model registration by collecting the information that has not been well employed in FP-based approaches.

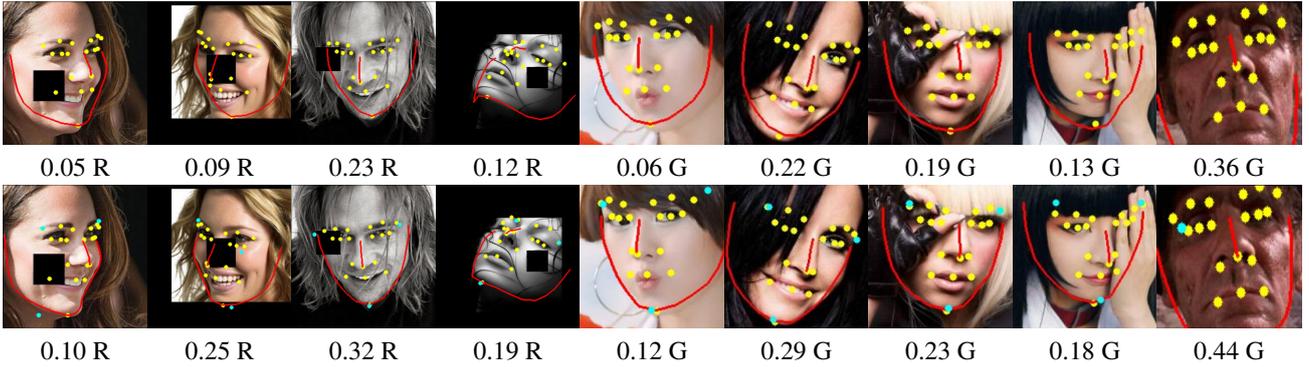

| 0.05 R | 0.09 R | 0.23 R | 0.12 R | 0.06 G | 0.22 G | 0.19 G | 0.13 G | 0.36 G |
| 0.10 R | 0.25 R | 0.32 R | 0.19 R | 0.12 G | 0.29 G | 0.23 G | 0.18 G | 0.44 G |

Figure 8: Re-projected FPs and silhouette outlines of 3D models after registration. Pose estimated by: (T) Detected FPs, (B) SHR. The generated alignment scores are shown below. The cyan dots indicate the final locations of SPs. The symbol 'R' indicates registered model is a reconstructed model, while 'G' indicates the model is generic. (Better viewed in color)

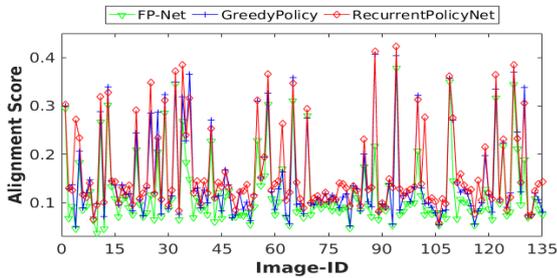

Figure 6: Alignment score under different policy. (Better viewed in color)

**Register a generic 3D model:** The objective of this experiment is to demonstrate that the proposed model registration approach can register a generic 3D model under occlusion. The COFW database is employed for evaluation. Since a personalized 3D model is unavailable, a generic 3D model is employed for registration. We observe the NRME is not a good criterion in this case because it is always affected by the inconsistency of 3D model. Instead, we measure how the registered 3D model could correct the predictions made by FP-Net. After employing SHR, the projected FPs from the generic model are used to replace the (predicted) occluded FPs; the results are shown in Table 2. We observed that the proposed method is able to improve the predictions made by the FP-Net, which reflects better registration accuracy.

Table 2: NME of 19 FPs. The error is measured in percentage (%). NME larger than (4%) is considered as a failure case. The value of $\omega$ equals to the bounding box size when computing the NME.

| Method | NME (All) | NME (Occ.) | Failure |
| --- | --- | --- | --- |
| RCPR [4] | 3.15 | 4.50 | 18.15 |
| FP-Net | 2.41 | 4.22 | 4.73 |
| FP-Net+SHR | **2.36** | **3.97** | **2.96** |

## 6. Conclusion

We proposed a new approach that is able to register a 3D facial model onto a 2D image under occlusion. The registration is based on the information encoded in the confidence-maps generated by fully convolutional networks. We introduced the concept of Sensiblepoints so that this information could be well exploited. Recurrent policy networks are employed to move the Sensiblepoints and seek a pose hypothesis that is able to maximize a non-differentiable alignment score. Experiments demonstrated that the proposed method is robust in registering reconstructed and generic facial 3D models under synthesized and natural occlusions.

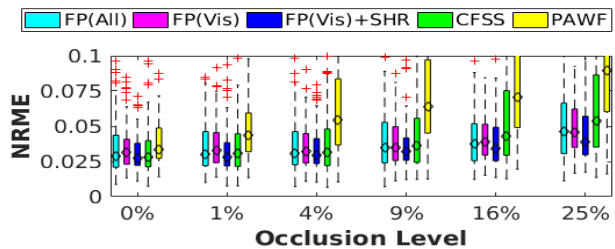

Figure 7: Boxplot of model registration error on iBUG-135 database with synthezised occlusions.

**Acknowledgement** This material is based upon work supported by the U.S. Department of Homeland Security under Grant Award Number 2015-ST-061-BSH001. This grant is awarded to the Borders, Trade, and Immigration (BTI) Institute: A DHS Center of Excellence led by the University of Houston, and includes support for the project "Image and Video Person Identification in an Operational Environment: Phase I" awarded to the University of Houston. The views and conclusions contained in this document are those of the authors and should not be interpreted as necessarily representing the official policies, either expressed or implied, of the U.S. Department of Homeland Security.

## References

[1] P. Belhumeur, D. Jacobs, D. Kriegman, and N. Kumar. Localizing parts of faces using a consensus of exemplars. In *Proc. IEEE Conference on Computer Vision and Pattern Recognition*, pages 545–552, Springs, CO, June 20-25 2011.

[2] A. Bulat and G. Tzimiropoulos. Convolutional aggregation of local evidence for large pose face alignment. In *Proc. British Machine Vision Conference*, York, UK, Sep. 9-22 2016.

[3] A. Bulat and G. Tzimiropoulos. How far are we from solving the 2D and 3D face alignment problem? (and a dataset of 230,000 3D facial landmarks). *arXiv preprint arXiv:1703.07332*, pages 1–10, 2017.

[4] X. P. Burgos-Artizzu, P. Perona, and P. Dollár. Robust face landmark estimation under occlusion. In *Proc. IEEE International Conference on Computer Vision*, pages 1–8, Sydney, Australia, Dec. 3-6 2013.

[5] D. F. Dementhon and L. S. Davis. Model-based object pose in 25 lines of coden. *International Journal of Computer Vision*, 15:123–141, 1995.

[6] P. Dou, Y. Wu, S. K. Shah, and I. A. Kakadiaris. Benchmarking 3D pose estimation for face recognition. In *Proc. IEEE International Conference on Pattern Recognition*, pages 190 – 195, Stockholm, Sweden, Aug. 24-28 2014.

[7] P. Dou, L. Zhang, Y. Wu, S. K. Shah, and I. A. Kakadiaris. Pose-robust face signature for multi-view face recognition. In *Proc. International Conference on Biometrics: Theory, Applications and Systems*, pages 1–8, Arlington, VA, Sep. 8-11 2015.

[8] M. Fischler and R. Bolles. Random sample consensus: A paradigm for model fitting with applications to image analysis and automated cartography. *Communications of the Association for Computing Machinery*, 24(6):381–395, 1981.

[9] G. Ghiasi and C. C. Fowlkes. Using segmentation to predict the absence of occluded parts. In *Proc. British Machine Vision Conference*, Swansea, UK, Sep. 7-10 2015.

[10] R. Hartley and A. Zisserman. *Multiple view geometry in computer vision*. Cambridge University Press, 2000.

[11] T. Hassner, S. Harel, E. Paz, and R. Enbar. Effective face frontalization in unconstrained images. In *Proc. IEEE Conference on Computer Vision and Pattern Recognition*, pages 4295 – 4304, Boston, Massachusetts, June 7 - 12 2015.

[12] T. Hassner, I. Masi, J. Kimand, J. Choi, S. Harel, P. Natarajan, and G. Medioni. Pooling faces: template based face recognition with pooled face image. In *Proc. IEEE Conference on Computer Vision and Pattern Recognition Workshops*, Las Vegas, NV, June 26-July 1 2016.

[13] S. Hochreiter and J. Schmidhuber. Long short-term memory. *Neural Computation*, 9(8):1735–1780, 1997.

[14] K. Jarvelin and J. Kekalainen. Cumulated gain-based evaluation of ir techniques. *ACM Transactions on Information Systems*, 20(4):422–446, 2002.

[15] A. Jourabloo and X. Liu. Large-pose face alignment via CNN-based dense 3D model fitting. In *Proc. IEEE Conference on Computer Vision and Pattern Recognition*, pages 4188 – 4196, Las Vegas, NV, June 26-July 1 2016.

[16] I. A. Kakadiaris, G. Toderici, G. Evangelopoulos, G. Passalis, X. Zhao, S. K. Shah, and T. Theoharis. 3D-2D face recognition with pose and illumination normalization. *Computer Vision and Image Understanding*, 154:137–151, 2017.

[17] D. P. Kingma and J. Ba. Adam: A method for stochastic optimization. In *Proc. International Conference on Learning Representations*, San Diego, CA, May 7-9 2015.

[18] M. Kostinger, P. Wohlhart, P. M. Roth, and H. Bischof. Annotated facial landmarks in the wild: A large-scale, real-world database for facial landmark localization. In *Proc. IEEE International Workshop on Benchmarking Facial Image Analysis Technologies*, Barcelona, Spain, Nov. 13 2011.

[19] A. Krull, E. Brachmann, F. Michel, M. Yang, S. Gumhold, and C. Rother. Learning analysis-by-synthesis for 6d pose estimation in RGB-D images. In *Proc. IEEE International Conference on Computer Vision*, Santiago, Chile, Dec. 13-16 2015.

[20] A. Krull, E. Brachmann, S. Nowozin, F. Michel, J. Shjotton, and C. Rother. PoseAgent: Budget-constrained 6D object pose estimation via reinforcement learning. *arXiv preprint arXiv:arXiv:1612.03779*, pages 1–9, 2017.

[21] A. Kumar and R. Chellappa. A convolution tree with deconvolution branches: Exploiting geometric relationships for single shot keypoint detection. *arXiv preprint arXiv:1704.01880*, 2017.

[22] L. A. Maas, Y. A. Hannun, and Y. A. Ng. Rectifier nonlinearities improve neural network acoustic models. In *Proc. International Conference on Machine Learning*, Atlanta, USA, June 16-21 2013.

[23] A. Martinez and A. Kak. PCA versus LDA. *IEEE Trans. Pattern Analysis and Machine Intelligence*, 23(2):228–233, 2001.

[24] I. Masi, A. Trn, T. Hassner, J. Leksut, and G. Medioni. Do we really need to collect millions of faces for effective face recognition? In *Proc. European Conference on Computer Vision*, pages 579–596, Amsterdam, The Netherlands, October 11-14 2016.

[25] A. Moeini, H. Moeini, and K. Faez. Unrestricted pose-invariant face recognition by sparse dictionary matrix. *Image and Vision Computing*, 36(0):9 – 22, 2015.

[26] R. Pascanu, T. Mikolov, and Y. Bengio. On the difficulty of training recurrent neural networks. In *Proc. International Conference on Machine Learning*, pages 1–12, Atlanta, GA, June 16-21 2013.

[27] P. Paysan, R. Knothe, B. Amberg, S. Romdhani, and T. Vetter. A 3D face model for pose and illumination invariant face recognition. In *Proc. 6$^{th}$ IEEE International Conference on Advanced Video and Signal Based Surveillance*, pages 296–301, Genoa, Italy, Sep. 2-4 2009.

[28] X. Peng, R. S. Feris, X. Wang, and D. N. Metaxs. A recurrent encoder-decoder for sequential face alignment. In *Proc. European Conference on Computer Vision*, pages 38–56, Amsterdam, Netherlands, October 11-14 2016.

[29] R. Ranjan, V. M. Patel, and R. Chellappa. HyperFace: A deep multi-task learning framework for face detection, landmark localization, pose estimation, and gender recognition. *arXiv preprint arXiv:1603.01249*, pages 1–13, 2016.

[30] C. Sagonas, G. Tzimiropoulos, S. Zafeiriou, and M. Pantic. 300 faces in-the-wild challenge: The first facial landmark localization challenge. In *Proc. IEEE International Conference on Computer Vision Workshops*, pages 397–403, Sydney, Australia, Dec. 1-8 2013.

[31] Y. Taigman, M. Yang, M. Ranzato, and L. Wolf. DeepFace: Closing the gap to human-level performance in face verification. In *Proc. IEEE Conference on Computer Vision and Pattern Recognition*, pages 1701 – 1708, Columbus, Ohio, June 24-27 2014.

[32] R. J. Williams. Simple statistical gradient-following algorithms for connectionist reinforcement learning. *Machine Learning*, 8:229 – 256, 1992.

[33] Y. Wu and Q. Ji. Robust facial landmark detection under signicant head poses and occlusion. In *Proc. IEEE International Conference on Computer Vision*, Santiago, Chile, Dec. 13-16 2015.

[34] Y. Wu, S. K. Shah, and I. A. Kakadiaris. Rendering or normalization? An analysis of the 3D-aided pose-invariant face recognition. In *Proc. IEEE International Conference on Identity, Security and Behavior Analysis*, pages 1–8, Sendai, Japan, Feb. 29-Mar. 2 2016.

[35] Y. Wu, S. K. Shah, and I. A. Kakadiaris. GoDP: globally optimized dual pathway system for facial landmark localization in-the-wild. *ArXiv eprint*, pages 1–16, 2017.

[36] Y. Wu, X. Xu, S. Shah, and I. A. Kakadiaris. Towards fitting a 3D dense facial model to a 2D image: A landmark-free approach. In *Proc. of International Conference on Biometrics: Theory, Applications and Systems*, Arlington, VA, Sep. 8-11 2015.

[37] S. Xiao, J. Feng, J. Xing, H. Lai, S. Yan, and A. Kassim. Robust facial landmark detection via recurrent attentive-refinement networks. In *Proc. European Conference on Computer Vision*, pages 57–72, Amsterdam, Netherlands, Oct. 11-14 2016.

[38] E. Zhou, H. Fan, Z. Cao, Y. Jiang, and Q. Yin. Extensive facial landmark localization with coarse-to-fine convolutional network cascade. In *Proc. IEEE International Conference on Computer Vision Workshops*, pages 386 – 391, Sydney, Australia, Dec. 2-8 2013.

[39] S. Zhu, C. Li, C. C. Loy, and X. Tang. Face alignment by coarse-to-fine shape searching. In *Proc. IEEE Conference on Computer Vision and Pattern Recognition*, pages 4998–5006, Boston, MA, June 7 - 12 2015.

[40] S. Zhu, C. Li, C. C. Loy, and X. Tang. Unconstrained face alignment via cascaded compositional learning. In *Proc. IEEE Conference on Computer Vision and Pattern Recognition*, Las Vegas, NV, June 26-July 1 2016.

[41] X. Zhu, Z. Lei, X. Liu, H. Shi, and S. Z. Li. Face alignment across large poses: A 3D solution. In *Proc. IEEE Conference on Computer Vision and Pattern Recognition*, pages 146 – 155, Las Vegas, NV, June 26-July 1 2016.

[42] X. Zhu, Z. Lei, J. Yan, D. Yi, and S. Z. Li. High-fidelity pose and expression normalization for face recognition in the wild. In *Proc. IEEE Conference on Computer Vision and Pattern Recognition*, pages 787–796, Boston, MA, June 7-12 2015.